\documentclass[letterpaper, 10pt, conference]{ieeeconf}

\IEEEoverridecommandlockouts

\usepackage{comment}
\usepackage{amsmath}

\usepackage{graphicx}
\usepackage{rotating}
\usepackage{float}
\usepackage{mathrsfs}
\usepackage{amssymb}
\usepackage{autobreak}
\usepackage{mathtools}
\usepackage{wrapfig}
\usepackage{bbm}
\usepackage{algorithm}
\usepackage{graphicx, subcaption}
\usepackage{makecell}
\usepackage{booktabs}
\usepackage[svgnames]{xcolor}
\usepackage[normalem]{ulem}
\usepackage{gensymb}

\usepackage[left=0.75in, right=0.75in, top=0.75in, bottom=0.78in]{geometry}
\usepackage[]{algpseudocode}

\usepackage{lipsum}
\usepackage{outlines}
\usepackage{blindtext}
\usepackage{multicol}
\usepackage{tikz}
\usetikzlibrary{shapes,backgrounds}
\usepackage{array}
\usepackage{wrapfig}
\usepackage{thmtools} 
\usepackage{thm-restate}
 
\usepackage{epsfig} 
\usepackage{amsmath}
\usepackage{algorithm}
\usepackage[]{algpseudocode}
\usepackage{tablefootnote}
\usepackage{multirow}
\usepackage{listings}

\def\ie{\emph{i.e.,}}
\def\eg{\emph{e.g.,}}
\def\etal{\emph{et al.}}

\usepackage{glossaries-extra}
\setabbreviationstyle[acronym]{long-short}
\glssetcategoryattribute{acronym}{nohyper}{true}

\newacronym{slam}{SLAM}{Simultaneous Localization and Mapping}
\newacronym{ba}{BA}{Bundle Adjustment}
\newacronym{sfm}{SfM}{Structure from Motion}
\newacronym{pgo}{PGO}{Pose-Graph Optimization}
\newacronym{vpr}{VPR}{Visual Place Recognition}
\newacronym{sgd}{SGD}{Stochastic Gradient Descent}
\newacronym{ils}{ILS}{Iterative Least-Squares}
\newacronym{gn}{GN}{Gauss-Newton}
\newacronym{lm}{LM}{Levenberg-Marquardt}
\newacronym{sdp}{SDP}{Semi-Definite Programming}
\newacronym{vo}{VO}{Visual Odometry}
\newacronym{vio}{VIO}{Visual-Inertial Odometry}
\newacronym{imu}{IMU}{Inertial Measurement Units}
\newacronym{pnp}{PnP}{Perspective-n-Point}
\newacronym{dof}{DoF}{Degrees of Freedom}
\newacronym{ar}{AR}{Augmented Reality}
\newacronym{sota}{SOTA}{state-of-the-art}
\newacronym{rpr}{RPR}{Relative Pose Regression}
\newacronym{tsdf}{TSDF}{Truncated Signed Distance Field}
\newacronym{esdf}{ESDF}{Euclidean Signed Distance Field}
\newacronym{gvd}{GVD}{Generalized Voronoi Diagram}
\newacronym{gnc}{GNC}{Graduated Non-Convexity}
\newacronym{gcnn}{GCNN}{Graph Convolutional Neural Network}
\newacronym{llm}{LLM}{Large Language Models}
\newacronym{vlm}{VLM}{Vision Language Model}

\usepackage{etoolbox}
\AtBeginEnvironment{algorithmic}{\everymath{\small}}

\def\sota{\gls{sota} }
\def\slam{\gls{slam} }
\def\imu{\gls{imu}}

\def\tsdf{\gls{tsdf}}
\def\esdf{\gls{esdf}}

\def\gcnn{\gls{gcnn} }
\def\llm{\gls{llm}}
\def\vlm{\gls{vlm}}

\definecolor{revisionColor}{HTML}{FF0000}
\definecolor{backcolour}{rgb}{0.95,0.95,0.92}

\lstdefinestyle{ieeeprompt}{
  basicstyle=\ttfamily\small,
  breaklines=true,
  columns=fixed
  columns=fullflexible,
  backgroundcolor=\color{backcolour}
}

\usepackage{amsopn}

\newcommand{\bv}{\mathbf{v}}

\newcommand{\bF}{\mathbf{F}}

\newcommand{\cS}{\mathcal{S}}

\newcommand{\cN}{\mathcal{N}}

\newcommand{\cR}{\mathcal{R}}
\newcommand{\cX}{\mathcal{X}}
\newcommand{\cI}{\mathcal{I}}

\newcommand{\cG}{\mathcal{G}}
\newcommand{\cB}{\mathcal{B}}
\newcommand{\cE}{\mathcal{E}}
\newcommand{\bbR}{\mathbb{R}}
\newcommand{\bbN}{\mathbb{N}}

\newcommand{\bb}{\mathbf{b}}
\newcommand{\bc}{\mathbf{c}}

\newcommand{\bx}{\mathbf{x}}

\newcommand{\bff}{\mathbf{f}}
\newcommand{\bp}{\mathbf{p}}

\def\g2o{$g^2o$}
\def\t2v{\mathrm{t2v}}
\def\v2t{\mathrm{v2t}}
\def\ev2t{\mathrm{ev2t}}

\def\auc{$\text{AUC}_{k}^{\text{Acc}}$}
\def\sr{$\text{SR}\%$}
\def\fp{FP}

\newcommand{\mypar}[1]{\noindent\textbf{#1}.}
\newcommand{\method}{\textsc{\small{ReasoningGraph}}}

\newcommand{\level}{L}
\newcommand{\mesh}{\level_1}
\newcommand{\object}{\level_2}
\newcommand{\place}{\level_3}
\newcommand{\room}{\level_4}
\newcommand{\building}{\level_5}
\newcommand{\numlayers}{N}

\newcommand{\rethree}{\bbR^{3}}

\newcommand{\resix}{\bbR^{6}}
\newcommand{\re}[1]{\bbR^{#1}}

\newcommand{\point}[2]{\bp_{#1}^{#2}}

\newcommand{\graph}{\cG}
\newcommand{\nodes}[1]{\cN_{#1}}
\newcommand{\edges}[2]{\cE_{#1}^{#2}}
\newcommand{\node}[1]{N_{#1}}
\newcommand{\edge}[2]{E_{#1}^{#2}}
\newcommand{\pointcolor}[2]{\bc_{#1}^{#2}}
\newcommand{\colorspace}{[0, 255]^3}
\newcommand{\semanticlabel}[1]{s_{#1}}
\newcommand{\naturals}{\bbN}
\newcommand{\vertex}[1]{\bv_{#1}}

\newcommand{\bbox}[1]{\bb_{#1}}
\newcommand{\feature}[2]{\bff_{#1}^{#2}}
\newcommand{\featurematrix}[2]{\bF_{#1}^{#2}}
\newcommand{\relationship}{r}
\newcommand{\image}[2]{I_{#1}^{#2}}
\newcommand{\pose}[2]{\bx_{#1}^{#2}}
\newcommand{\bboxestwod}[2]{\cB_{#1}^{#2}}
\newcommand{\task}[1]{\textbf{T#1}}
\newcommand{\relations}{\cR}
\newcommand\boldblue[1]{\textcolor{RoyalBlue}{\textbf{#1}}}
\newcommand\boldgreen[1]{\textcolor{Green}{\textbf{#1}}}

\title{\LARGE \bf Relationship-Aware Hierarchical 3D Scene Graph for Task Reasoning}

\author{Albert Gassol Puigjaner, Angelos Zacharia, Kostas Alexis
\thanks{Autonomous Robots Lab, Norwegian University of Science and Technology (NTNU), Trondheim, Norway,
    {\tt \footnotesize
        \href{mailto:albert.g.puigjaner@ntnu.no}{albert.g.puigjaner@ntnu.no}}
    }
    \thanks{This work was supported by the European Commission Horizon Europe grant SYNERGISE (EC 101121321).}
}

\definecolor{lightblue}{rgb}{0.12,0.49,0.85}
\usepackage[colorlinks=true,linkcolor=black,citecolor=blue,urlcolor=blue]{hyperref}
\usepackage[capitalise]{cleveref}

\begin{document}

\maketitle

\begin{abstract}
Representing and understanding 3D environments in a structured manner is crucial for autonomous agents to navigate and reason about their surroundings. While traditional \slam methods generate metric reconstructions and can be extended to metric-semantic mapping, they lack a higher level of abstraction and relational reasoning. To address this gap, 3D scene graphs have emerged as a powerful representation for capturing hierarchical structures and object relationships. In this work, we propose an enhanced hierarchical 3D scene graph that integrates open-vocabulary features across multiple abstraction levels and supports object-relational reasoning. Our approach leverages a \vlm~to infer semantic relationships. Notably, we introduce a task reasoning module that combines \llm~and a \vlm~to interpret the scene graph’s semantic and relational information, enabling agents to reason about tasks and interact with their environment more intelligently. We validate our method by deploying it on a quadruped robot in multiple environments and tasks, highlighting its ability to reason about them.
\end{abstract}

\setlength{\textfloatsep}{5pt}

\glsreset{slam}
\glsreset{llm}
\glsreset{vlm}

\section{INTRODUCTION}\label{sec:intro}
A central challenge in spatial perception for robotics is constructing 3D representations that are both structured and semantically meaningful. Humans naturally perceive and manipulate scenes by recognizing objects, their properties, and relationships, including hierarchical structures such as rooms within floors or buildings. For autonomous agents, this requires scalable, online representations that support multiple levels of abstraction and reasoning about object relationships.

Traditional \gls{slam} methods reconstruct metric maps from sensors like cameras~\cite{schoenberger2016mvs}, LiDARs~\cite{Khedekar2022Mimosa}, radars~\cite{Nissov2024DegradationRL}, or \imu~\cite{Forster2017Preintegration}, and can be extended to closed-~\cite{Rosinol2021Kimera} or open-vocabulary~\cite{Peng2023OpenScene} metric-semantic maps using vision foundation models~\cite{Radford2021CLIP,Kirillov2023SAM}. However, these approaches lack a higher level of abstraction and object reasoning.

3D scene graphs~\cite{Rosinol2021Kimera,armeni20193dscenegraphs,hughes2022hydra,hughes2024foundations,Wals20203DSSG,Wu2021SceneGraphFusion} address this gap by capturing hierarchical and relational information. Hierarchical models~\cite{armeni20193dscenegraphs,hughes2022hydra,hughes2024foundations} represent indoor scenes at multiple abstraction levels (\eg~objects, rooms, buildings) while also encoding geometrical inter- and intra-layer relationships (\eg~an object being inside a room). Other works~\cite{Wals20203DSSG,Wu2021SceneGraphFusion} generate object-level scene graphs from RGB point clouds, predicting geometric, comparative, and semantic relationships, while recent methods~\cite{koch2024open3dsg} leverage a \gls{vlm}~to introduce open-vocabulary object relationships. More recently, incremental approaches~\cite{Gu2024conceptgraphs,werby2023hovsg} 
construct open-vocabulary scene graphs, without explicit object-level relationships.

\begin{figure}[t]
    \centering
    \includegraphics[width=\columnwidth]{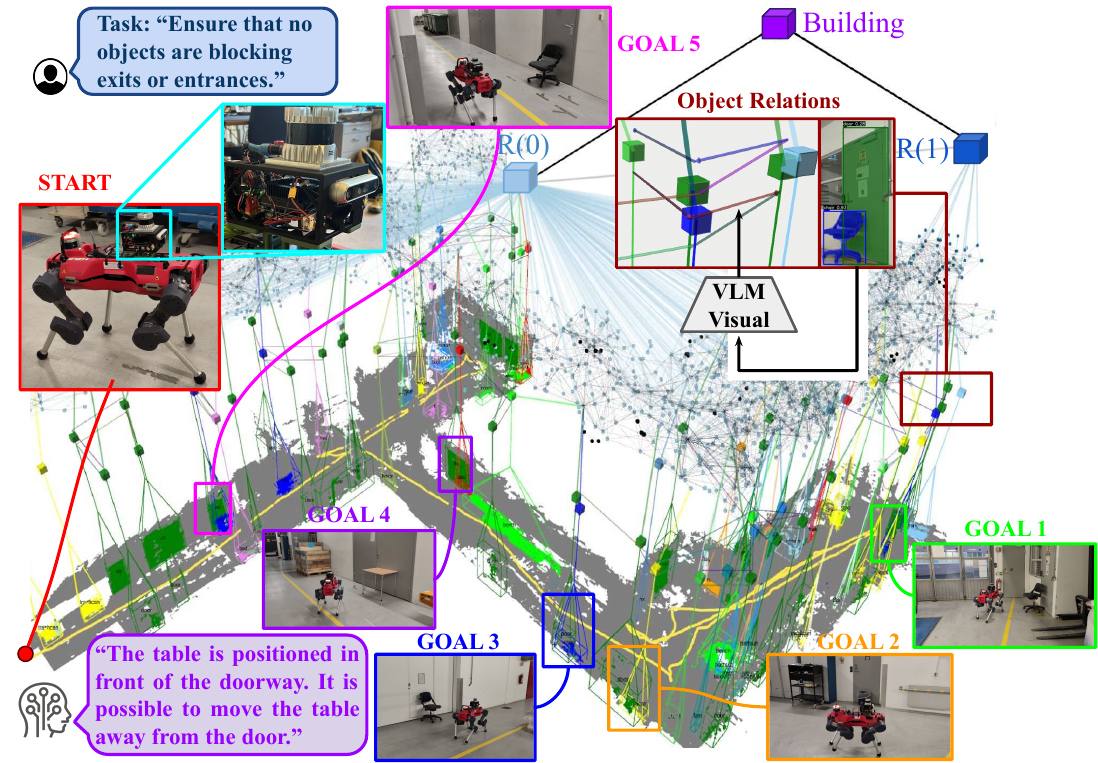}
    \caption{ Task reasoning example. We deploy \method~on a quadruped robot, which incrementally builds an open-vocabulary, relationship-aware hierarchical scene graph of the environment during autonomous exploration. Leveraging open-vocabulary and object-relational embeddings, \method~identifies task-relevant objects and reasons about their interactions. In this example, it identifies all the objects (chairs, a table, and a trash can) that are blocking the exits.}
\label{fig:example}
\vspace{-0.6em}
\end{figure}

In this work, we propose \method, a framework for incrementally constructing a reasoning-enhanced hierarchical 3D scene graph that integrates open-vocabulary features across multiple levels of abstraction and supports object-relational reasoning. Additionally, we introduce a task reasoning module that, given a task that may require object-interaction reasoning (\eg~``prepare the room for a meeting'', ``ensure that exits are not blocked''), leverages the semantic and relational information in our graph to decompose the task into subtasks, identify the relevant objects, and evaluate which subtasks need to be executed. Our contributions are:

\begin{itemize}
    \item We extend hierarchical 3D scene graphs with open-vocabulary features across multiple abstraction levels.
    \item We leverage a \vlm~to infer object relationship features for a richer context-aware representation.
    \item  We propose a reasoning module that combines \llm~and a \vlm~to process natural language tasks, predict relevant objects, and assess object-interaction feasibility.
    \item We quantitatively evaluate \method’s ability to encode open-vocabulary objects, showing competitive performance against strong baselines. 
    \item We demonstrate the benefit of incorporating object relations together with our reasoning module to reason about complex tasks. Furthermore, we deploy our method on a quadruped robot, demonstrating its ability to build the scene graph online and reason about tasks. 
\end{itemize}

In the remainder of this paper, we first review related literature (\cref{sec:related}) and then define the problem addressed (\cref{sec:problem}). We proceed with a detailed description of the proposed method (\cref{sec:method}), followed by its performance evaluation (\cref{sec:results}). Conclusions are drawn in~\cref{sec:conclusions}.


\section{RELATED WORK}\label{sec:related}

\mypar{Metric-Semantic Representations} Closed-vocabulary semantic \slam methods aim to build semantically annotated 3D maps of the environment. These approaches typically rely on semantic and panoptic segmentation networks~\cite{Zhang2024EfficientViTSAMAS,Li2023MaskDINO} to enhance the 3D representation~\cite{Rosinol2021Kimera,Narita2019PanopticFusion}. With the introduction of vision foundation models such as CLIP~\cite{Radford2021CLIP} and SAM~\cite{Kirillov2023SAM}, recent works focus on constructing 3D representations enriched with open-vocabulary features that can be easily queried and/or clustered~\cite{Peng2023OpenScene}. These methods extract open-vocabulary embeddings from 2D images using vision foundation models and project them into 3D space. However, because they assign features at the point level, they require significant memory and do not scale efficiently. Additionally, these methods lack abstraction, hierarchical structure, and object-level reasoning, limiting their ability to support a higher level of scene understanding.

\mypar{3D Scene Graphs}Early works~\cite{Rosinol2021Kimera,armeni20193dscenegraphs} introduced 3D scene graphs to model indoor multi-level abstractions, enabling spatial reasoning across agent poses, objects, rooms, and buildings. These methods also established inter- and intra-layer relationships, such as object containment and spatial proximity, to provide a richer structural understanding of the scene. Subsequent works~\cite{hughes2022hydra,hughes2024foundations} adopted this hierarchical framework and extended it to incrementally build 3D scene graphs in real time. Meanwhile, object-level scene graphs~\cite{Wals20203DSSG,Wu2021SceneGraphFusion} have been proposed to infer geometric, comparative, and semantic relationships from RGB point clouds, further improving contextual reasoning. More recently, \vlm s have been leveraged to introduce open-vocabulary relationships, enabling flexible and adaptable semantics in scene graphs. Koch \etal~\cite{koch2024open3dsg} propose to distill the knowledge of the visual encoder of InstructBLIP~\cite{dai2023instructblip} into a \gcnn, while Chen~\etal~\cite{Chen2024CLIPDriven} adopt a similar approach with CLIP~\cite{Radford2021CLIP}. However, these methods typically construct object-level scene graphs offline, requiring a complete point cloud of the scene. Despite these advancements, existing approaches still suffer from several limitations. Namely, they either (a) lack open-vocabulary and relationship reasoning, (b) fail to incorporate hierarchical representations, or (c) are unsuitable for online scene graph construction.

\mypar{Open-Vocabulary 3D Scene Graphs} Recent works~\cite{Gu2024conceptgraphs,werby2023hovsg} have explored open-vocabulary scene graphs to enable language-grounded navigation. ConceptGraphs~\cite{Gu2024conceptgraphs} incrementally constructs object-level open-vocabulary 3D scene graphs by clustering 3D-projected CLIP embeddings. It further predicts object relationships by prompting GPT-4 with geometric information and summarized image captions of objects. Additionally, it enables language-grounded object search by querying GPT-4 with the objects' geometric data and captions. However, this approach is limited to small-scale scenes and lacks a hierarchical structure in its representation. 

Building upon similar ideas, HOV-SG~\cite{werby2023hovsg} introduces a hierarchical open-vocabulary 3D scene graph, organized into building, floors, rooms, objects, and a navigational graph. In addition to attaching CLIP embeddings to detected objects, where object embeddings are projected into 3D using depth images and averaged across multiple views, HOV-SG also extends open-vocabulary features to floors and rooms within its hierarchy. During object search, GPT-4 parses natural language queries into structured attributes: the room, floor, and object mentioned in the query. These attributes are then matched within the hierarchy by computing cosine similarities with the open-vocabulary embeddings. While this method integrates both hierarchical and open-vocabulary representations, it lacks relationship reasoning, which could further improve language-grounded task reasoning by capturing relevant interactions between objects.

\section{PROBLEM STATEMENT} 
\label{sec:problem}

\begin{figure*}[t]
    \centering
    \includegraphics[width=\linewidth]{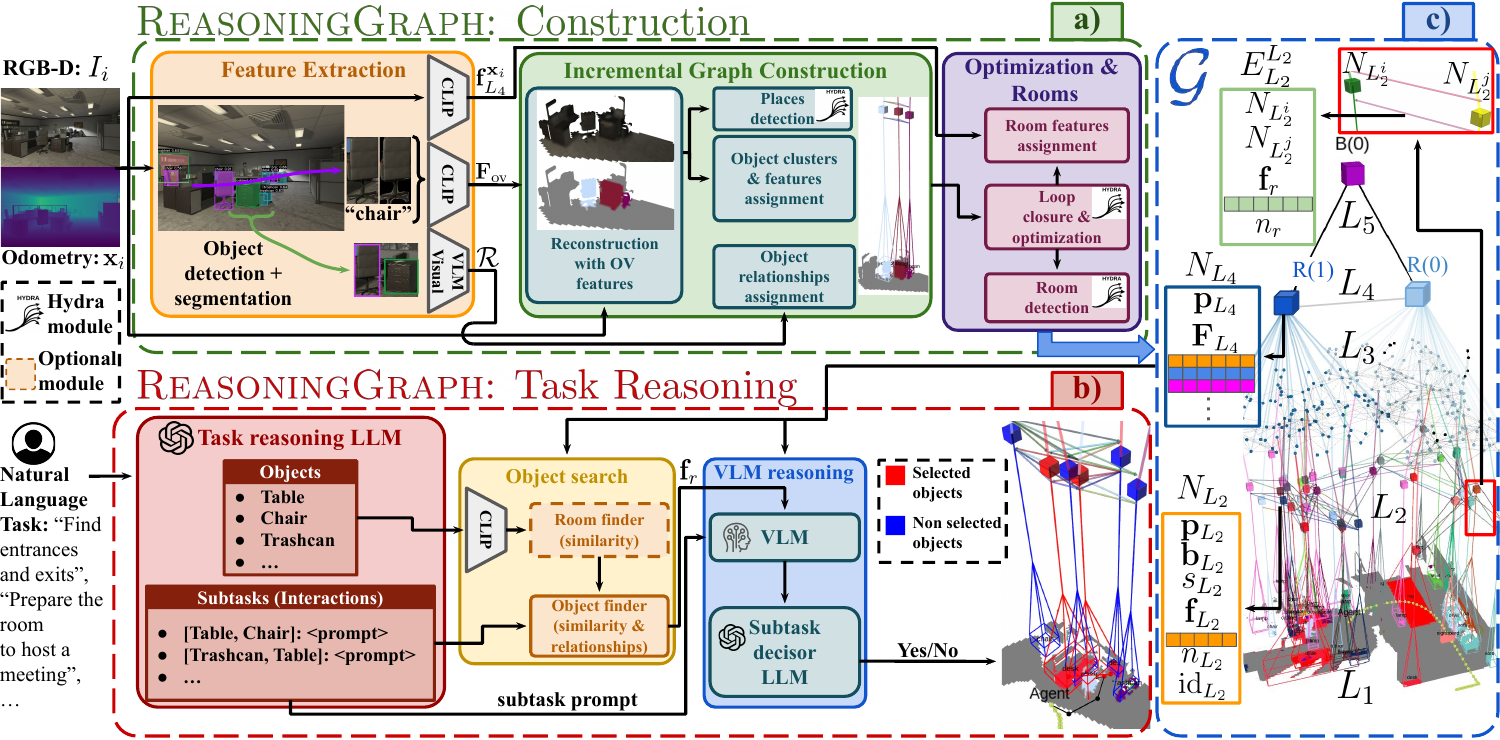}
    \caption{\method~overview.\textbf{ a)} \method~incrementally builds a hierarchical 3D scene graph $\color{RoyalBlue}\graph$ (\textbf{c)}) from RGB-D frames and poses, using an open-vocabulary detector~\cite{wang2025yoloe} and CLIP~\cite{Radford2021CLIP} embeddings for object representation. Object relations are derived from a \vlm~\cite{Lu2024DeepSeekVLTR} visual encoder, while Hydra~\cite{hughes2022hydra} reconstructs the semantic mesh ($\mesh$), clusters objects ($\object$), and detects places and rooms ($\place$, $\room$). Open-vocabulary features and relations are then assigned to $\color{RoyalBlue}\graph$.
    \textbf{b)} The task reasoning module leverages two \llm s and a \vlm. Given a task, the \llm~identifies relevant objects and formulates subtasks needing evaluation. These subtasks are evaluated for feasibility by the \vlm, with CLIP similarity used for object retrieval.}
    \label{fig:overview}
    \vspace{-2em}
\end{figure*}

A hierarchical 3D scene graph $\graph = \langle \nodes{}, \edges{}{} \rangle$, consisting of $\numlayers$ hierarchical layers, is defined by a set of nodes $\nodes{} = \{\nodes{\level_n}\}_{n=1}^{\numlayers}$, which contains all nodes across the layers, and a set of edges $\edges{}{}$ representing geometric or semantic relationships between them. Layers are organized from bottom to top, with each successive layer representing entities at a higher level of abstraction. Each node has geometric properties, such as position or centroid, and may include semantic and/or open-vocabulary attributes. Each layer $\level_n$ includes intra-layer edges $\edges{\level_n}{\level_n}$, which encode relational information. Additionally, inter-layer edges $\edges{L_n}{L_{n+1}}$ connect nodes between layers. Formally, the sets of nodes and edges of the graph are defined as:
\begin{align}
    \nodes{} &= \bigcup_{n = 1}^{\numlayers} \nodes{\level_n}, \\
    \edges{}{} &= \left( \bigcup_{n = 1}^{\numlayers - 1} \left( \edges{\level_n}{\level_n} \cup \edges{\level_n}{\level_{n+1}} \right) \right) \cup \edges{\level_{\numlayers}}{\level_{\numlayers}}.
\end{align}

\noindent
\textit{Objective 1}: Given a sequence of $M \in \naturals$ RGB-D frames $\cI = \{\image{i}{}\}_{i=1}^{M}$, where each frame $\image{i}{} = \{\image{i}{\text{RGB}}, \image{i}{\text{Depth}}\}$, along with corresponding odometry estimates $\cX = \{\pose{i}{}\}_{i=1}^{M}$ of an indoor scene, the goal is to incrementally construct $\graph$. Additionally,  we aim to progressively enrich $\graph$ with open-vocabulary semantics and relational information.

 \noindent
 \textit{Objective 2}: Given the scene graph $\graph$, the objective is to leverage its open-vocabulary and relational representation to reason about tasks involving object interaction or search, by identifying relevant objects and determining whether interactions among them are needed to accomplish the task.


\section{METHOD}\label{sec:method}

In this section, we briefly introduce our scene graph definition and its construction in~\cref{subsec:graph_def}, followed by a detailed discussion of open-vocabulary features and relations in~\cref{subsec:open_vocab_reasoning}. Finally, we present our reasoning module in~\cref{subsec:reasoning_module}, which leverages the scene graph information to reason about a given task. The overall design of our method is illustrated in~\cref{fig:overview}.

\vspace{-0.5em}
\subsection{Hierarchical Scene Graph Definition and Construction}\label{subsec:graph_def}

Following Hydra~\cite{hughes2022hydra,hughes2024foundations}, we define a five-layer hierarchy ($\numlayers=5$), illustrated in~\cref{fig:overview}c: 
\begin{itemize}
    \item \textbf{Metric-Semantic Mesh Layer} $\mesh$: Each node is $\node{\mesh} = \{\vertex{}, \pointcolor{}{}, \semanticlabel{}\} \in \nodes{\mesh}$, where $\vertex{} \in \rethree$ is a mesh vertex, $\pointcolor{}{} \in \colorspace$ its color, and $\semanticlabel{} \in \naturals$ its semantic label. If a vertex belongs to an object, we add a graph edge $\edge{\mesh}{\object} = \{\node{\mesh}, \node{\object}\} \in \edges{\mesh}{\object}$.
    
    \item  \textbf{Object Layer} $\object$:  As shown in~\cref{fig:overview}c, each object node is $\node{\object} = \{\point{\object}{}, \bbox{\object}{}, \semanticlabel{\object}, \feature{\object}{}, n_{\object}, \text{id}_{\object}\} \in \nodes{\object}$, where $\point{\object}{} \in \rethree$ is the object centroid, $\bbox{\object}{} \in \resix$ its bounding box, $\semanticlabel{\object} \in \naturals$ its label, $\feature{\object}{} \in \re{|\feature{\object}{}|}$ an open-vocabulary feature, $n_{\object} \in \naturals$ the feature update count, and $\text{id}_{\object} \in \naturals$ a variable identifier used for object-level relationships assignment. Objects connect to their nearest place via $\edge{\object}{\place} = \{\node{\object}, \node{\place}\} \in \edges{\object}{\place}$. Object-level semantic relations (\eg~``a glass is on a table'') between two objects ($\node{\object^i}$ and $\node{\object^j}$) are captured with $\edge{\object^i}{\object^j} = \{\node{\object^i}, \node{\object^j}, \feature{\relationship}{}, n_{\relationship}\} \in \edges{\object}{\object}$, where $\feature{\relationship}{} \in \re{|\feature{\relationship}{}|}$ encodes the relation and $n_{\relationship} \in \naturals$ counts its updates. Such relations can be visualized in~\cref{fig:overview}c.
    
    \item \textbf{Place Layer} $\place$: Each node is $\node{\place} = \{\point{\place}{}\} \in \nodes{\place}$, with $\point{\place}{} \in \rethree$ as centroid. Places connect to their rooms via the graph edge $\edge{\place}{\room} = \{\node{\place}, \node{\room}\} \in \edges{\place}{\room}$.
    
    \item \textbf{Room Layer} $\room$:
    Each node is $\node{\room} = \{\point{\room}{}, \featurematrix{\room}{}\} \in \nodes{\room}$, where $\point{\room}{} \in \rethree$ is the room centroid and $\featurematrix{\room}{} = [\feature{\room}{1}, \dots, \feature{\room}{K}]^T \in \re{K \times |\feature{\room}{}|}$ a set of $K$ open-vocabulary feature clusters, with $\feature{\room}{i} \in \re{|\feature{\room}{}|}$ being the $i$-th open-vocabulary feature cluster of the set (see~\cref{fig:overview}c for a visual representation of room nodes). Rooms connect to buildings via the edge $\edge{\room}{\building} = \{\node{\room}, \node{\building}\} \in \edges{\room}{\building}$.
    
    \item \textbf{Building Layer} $\building$: 
    Each node is $\node{\building} = \{\point{\building}{}\} \in \nodes{\building}$, with $\point{\building}{} \in \rethree$ as centroid.
\end{itemize}

Having defined the hierarchical scene graph, we now describe its construction (\cref{fig:overview}a) from sensor data. We employ Hydra~\cite{hughes2022hydra} to reconstruct the scene and extract the hierarchical graph layers ($\mesh \text{ - } \building$). Hydra incrementally builds the mesh ($\mesh$), object ($\object$), and place ($\place$) layers online, while the room layer ($\room$), mesh refinement, and pose optimization are updated at a lower frequency. This produces the hierarchical structure, which serves as the basis for open-vocabulary features and relational reasoning enhancement.

The semantic mesh ($\nodes{\mesh}$) is constructed using Kimera~\cite{Rosinol2021Kimera} for semantic segmentation and a windowed Voxblox~\cite{Oleynikova2017Voxblox} to extract the \tsdf, \esdf, and mesh via marching cubes. 

Objects ($\nodes{\object}$) are extracted via Euclidean clustering of vertices with the same semantic label, and overlapping objects of the same class are incrementally fused. Places ($\nodes{\place}$) are obtained by sparsifying a Generalized Voronoi Diagram (GVD) derived from the ESDF, connecting the resulting voxels to form a graph. Rooms ($\nodes{\room}$) are detected by dilating the voxel map and pruning the corresponding subgraph of places, such that connected place nodes correspond to rooms.

\vspace{-0.5em}
\subsection{Open-Vocabulary and Reasoning Enhancement}\label{subsec:open_vocab_reasoning}

We enhance the graph $\graph$ by attaching open-vocabulary features to objects ($\nodes{\object}$) and rooms ($\nodes{\room}$). Furthermore, we compute relational features between objects and incorporate them as graph edges ($\edges{\object}{\object}$). This enhancement can be visualized in~\cref{fig:overview}c. Introducing such features requires several steps in our framework. We present these in~\cref{alg:open_vocab_low_level,alg:open_vocab_mid_level,alg:open_vocab_rels_assignment,alg:open_vocab_high_level}. Next, we provide details on how these features are computed and included in $\graph$.

\mypar{Object Features} In~\cref{alg:open_vocab_low_level} (\cref{low_level:segmentation}), we begin by detecting object bounding boxes, segmentation masks and semantic labels ($\bboxestwod{}{}$, $\image{}{\text{seg}},\cS$) from the input RGB-D frame $\image{}{} = \{ \image{}{\text{RGB}}, \image{}{\text{Depth}} \}$ (frame index dropped for simplicity), using an open-set detection and segmentation method such as YOLOe~\cite{wang2025yoloe}. For each detected object $i \in \{1,\ldots,|\bboxestwod{}{}|\}$, we generate two image crops: (a) a masked image $g_{\text{mask}}(\image{}{\text{RGB}}, \image{i}{\text{seg}})$, where the object is isolated with a black background, and (b) a bounding-box crop $g_{\bboxestwod{}{}}(\image{}{\text{RGB}}, \bboxestwod{i}{})$.
We then get averaged CLIP~\cite{Radford2021CLIP} embeddings:
\begin{align}
\label{eq:average_clip}
\feature{\text{ov}}{i} = \, &\alpha_{\text{mask}}\feature{\text{CLIP}}{}(g_{\text{mask}}(\image{}{\text{RGB}}, \image{i}{\text{seg}})) + \\ \notag
&\alpha_{\bboxestwod{}{}}\feature{\text{CLIP}}{}(g_{\bboxestwod{}{}}(\image{}{\text{RGB}}, \bboxestwod{i}{})) + \alpha_{\semanticlabel{}}\feature{\text{CLIP}}{}(\cS_i),
\end{align}
where $\feature{\text{CLIP}}{}(g_{\text{mask}}(\image{}{\text{RGB}}, \image{i}{\text{seg}}))$ is the embedding of the masked object, $\feature{\text{CLIP}}{}(g_{\bboxestwod{}{}}(\image{}{\text{RGB}}, \bboxestwod{i}{}))$ is the embedding of the cropped object defined by its bounding box and $\feature{\text{CLIP}}{}(\cS_i)$ is the embedding of the object’s semantic label. 
Following~\cite{werby2023hovsg}, we combine both cropped and masked embeddings since this enhances the robustness of the CLIP representation, while the semantic label embedding adds a complementary textual cue. 
The weights satisfy $\alpha_{\text{mask}} + \alpha_{\bboxestwod{}{}} + \alpha_{\semanticlabel{}} = 1$. Finally, the open-vocabulary features of all objects are collected into a vector of feature vectors, denoted as $\bF_{\text{ov}}$ (\cref{low_level:clip_stack}). This process is illustrated in the Feature Extraction block of~\cref{fig:overview}a.

Consequently, in~\cref{alg:open_vocab_mid_level}, object features are temporarily attached to the mesh ($\nodes{\mesh}$) when performing the 3D reconstruction with Voxblox~\cite{Oleynikova2017Voxblox} and marching cubes (\cref{mid_level:meshing}). For each object $C^i$ in the clustered mesh (\cref{mid_level:cluster}), with $i \in \{1,\ldots, N_C\}$ and $N_C \in \naturals$ being the number of clusters, we determine if its vertices contain open-vocabulary features. When they do, we average them to get $\feature{\text{mesh}}{i}$ (\cref{mid_level:average_mesh_feats}). If $C^i$ corresponds to a node of our graph ($\node{\object^i }$), we average its open-vocabulary feature (\cref{mid_level:running_average_feats,mid_level:n_plus_1}):
\begin{align}
    \feature{\object^i}{} = \dfrac{n_{\object^i}  \feature{\object^i}{} + \feature{\text{mesh}}{i}}{n_{\object^i} + 1}, \quad \feature{\object^i}{}, n_{\object^i} \in \node{\object^i}.
    \label{eq:average_node_feature}
\end{align}

\vspace{-0.5em}
Otherwise, we create a new $\node{\object}$ node from the cluster $C^i$ and add it to the graph (\cref{mid_level:node_to_graph}). The detected open-vocabulary features are then removed from the mesh ($\nodes{\mesh}$) to reduce memory usage (\cref{mid_level:clean_mesh}).

  \begin{algorithm}[!tbp]
  \caption{Features Extraction}\label{alg:open_vocab_low_level}
    \begin{algorithmic}[1]
      \State \textbf{Input:} $\pose{}{}$, $\image{}{}$\phantomsection\label{low_level:input}
      \State $\bboxestwod{}{},\image{}{\text{seg}}, \cS = \texttt{detect}(\image{}{})$ \Comment{Object detection}\phantomsection\label{low_level:segmentation}
      \For{$i=1,\ldots,|\bboxestwod{}{}|$} \Comment{Average CLIP for each object}\phantomsection\label{low_level:for_clip}
        \State   $\feature{\text{ov}}{i} = \texttt{average}_{\text{CLIP}}(\image{}{\text{RGB}}, \image{i}{\text{seg}}, \bboxestwod{i}{}, \cS_i)$\phantomsection\label{low_level:clip}  \Comment{\cref{eq:average_clip}}
      \EndFor\phantomsection\label{low_level:end_clip}
      \State $\bF_{\text{ov}} = \{ \feature{\text{ov}}{0}, \ldots, \feature{\text{ov}}{|\bboxestwod{}{}|}  \}$\phantomsection\label{low_level:clip_stack}
      \State $\feature{\room}{\pose{}{}} = \feature{\text{CLIP}}{}(\image{}{\text{RGB}})$ \Comment{CLIP of the full input for rooms}\phantomsection\label{low_level:full_clip}
      \State $\relations = \{(i,j) : \feature{\text{VLM}}{}(\image{}{\text{RGB}}[\bboxestwod{i}{} \cup \bboxestwod{j}{}] )\} \quad \forall i,j \in \{1,\ldots,|\bboxestwod{}{}|\}, i \neq j$ \quad \Comment{\vlm~visual encoder to get object-relation features. Stored in a dictionary of object pairs}\phantomsection\label{low_level:ids_relations_map}
      \State \Return $\image{}{\text{seg}}$,  $\cS $, $\bF_{\text{ov}}$, $\feature{\room}{\pose{}{}}$, $\relations$
    \end{algorithmic}
  \end{algorithm}
  
  \begin{algorithm}[h]
  \caption{Object Features Assignment}\label{alg:open_vocab_mid_level}
    \begin{algorithmic}[1]
     \State \textbf{Input:} $\pose{}{}$, $\image{}{}$, $\image{}{\text{seg}}$, $\cS$, $\bF_{\text{ov}}$, $\relations$\phantomsection\label{mid_level:input}
      \State $\nodes{\mesh} = \texttt{reconstruct}_{\text{ov}}(\nodes{\mesh},\pose{}{}, \image{}{}, \image{}{\text{seg}}, \bF_{\text{ov}}, \cS, \relations)$ \phantomsection\label{mid_level:meshing}
      \State $C = \{C^1, \ldots, C^{N_{C}} \} = \texttt{cluster}(\nodes{\mesh})$\phantomsection\label{mid_level:cluster}
      \For{$i = 1, \ldots, N_{C}$} \Comment{Iterate over object clusters}\phantomsection\label{mid_level:for_ov}
          \State $\feature{\text{mesh}}{i} = \emptyset$, $n = 0$ \Comment{Empty feature}
          \If{$\nodes{\mesh}\texttt{.has\_features}(C^i)$}\Comment{Fill feature} \phantomsection\label{mid_level:features_in_mesh}
            \State $\feature{\text{mesh}}{i} = \texttt{average}(\nodes{\mesh}\texttt{.features}(C^i))$, $n = 1$ \phantomsection\label{mid_level:average_mesh_feats}
          \EndIf \phantomsection\label{mid_level:end_features_in_mesh}
          \If{$\node{\object^i} \in \nodes{\object}$} \Comment{Running average of features}\phantomsection\label{mid_level:node_in}
            \State $\feature{\object^i}{} = \texttt{average}_{\object}(\node{\object^i}, \feature{\text{mesh}}{i})$ \Comment{\cref{eq:average_node_feature}} \phantomsection\label{mid_level:running_average_feats}
            \State $n_{\object^i} \mathrel{+}= 1$,  $\text{id}_{\object^i} = \object^i.\text{id}$\phantomsection\label{mid_level:n_plus_1}
          \Else \Comment{Add new object node to the graph} \phantomsection\label{mid_level:else}
            \State $\nodes{\object} \gets \{ C^i.\point{\object}{}, C^i.\bbox{\object}{}, C^i.\semanticlabel{\object}, \feature{\text{mesh}}{i}, n, C^i.\text{id}\}$\phantomsection\label{mid_level:node_to_graph}
          \EndIf
      \EndFor\phantomsection\label{mid_level:end_ov}
      \State $\nodes{\mesh}\texttt{.remove}( \bF_{\text{ov}})$ \Comment{Delete mesh features}\phantomsection\label{mid_level:clean_mesh}
    \end{algorithmic}
  \end{algorithm}
  
  \begin{algorithm}[t]
  \caption{Room Features Assignment}\label{alg:open_vocab_high_level}
    \begin{algorithmic}[1]
      \State \textbf{Input:} $\{\feature{\room}{\pose{0}{}}, \ldots, \feature{\room}{\pose{n}{}}\}$, $\{\pose{0}{},\ldots,\pose{n}{}\}$\phantomsection\label{high_level:input}
      \State $\texttt{room\_features} = \{\}$
      \State $\nodes{\room} = \texttt{detect\_rooms}(\graph)$\phantomsection\label{high_level:detect_rooms}
      \For{$i=0, \ldots, n$}
        \State $\node{\room} = \texttt{find\_room}(\nodes{\room}, \pose{i}{})$ \Comment{Room containing $\pose{i}{}$}\phantomsection\label{high_level:find_room}
        \State $\texttt{room\_features}[\node{\room}]\texttt{.append}(\feature{\room}{\pose{i}{}})$\phantomsection\label{high_level:append_feature}
      \EndFor
      \For{$i=0, \ldots, |\nodes{\room}|$}\Comment{Room-feature clusters}
        \State $\featurematrix{\room^i}{} = \texttt{KMeans}(\texttt{room\_features}[\node{\room^i}])$\phantomsection\label{high_level:cluster}
      \EndFor
    \end{algorithmic}
  \end{algorithm}
  
  \begin{algorithm}[ht]
  \caption{Object Relationships Assignment}\label{alg:open_vocab_rels_assignment}
    \begin{algorithmic}[1]
    \State \textbf{Input:} $\relations$
    \For{$i = 1, \ldots, |\nodes{\object}| ; j = 1, \ldots, |\nodes{\object}|; j \neq i$}
          \State $\feature{\text{VLM}}{i,j} = \relations[(\text{id}_{\object^i}, \text{id}_{\object^j})]$\phantomsection\label{rels:get_relation}
          \State \textbf{if} $\feature{\text{{VLM}}}{i,j} == \emptyset$ \textbf{: continue} \Comment{Add/Update relations}\phantomsection\label{rels:ids_in_mesh}
          \If{$\edge{\object^i}{\object^j} \in \edges{\object}{\object}$} \Comment{Update existing relation}\phantomsection\label{rels:edge_in_graph}
          \State $\feature{\relationship}{i,j} = \texttt{average}_{\relationship}(\edge{\object^i}{\object^j}, \feature{\text{VLM}}{i,j})$ \Comment{\cref{eq:average_relation}} \phantomsection\label{rels:runing_average_reasoning}
          \State $n_{\relationship}^{i,j} \mathrel{+}= 1$ \phantomsection\label{rels:n_plus_1_relationship}
          \Else \Comment{Add new relationship to the graph}
              \State $\edges{\object}{\object} \gets \{ \node{\object^i}, \node{\object^j}, \feature{\text{VLM}}{i,j}, 1\}$ \phantomsection\label{rels:edge_to_graph}
          \EndIf
      \EndFor
      \State $\nodes{\mesh}\texttt{.remove}(\relations\texttt{.keys()})$ \Comment{Delete relation IDs}\phantomsection\label{rels_assignment:clean_mesh}
    \end{algorithmic}
  \end{algorithm}

\mypar{Rooms Features} Similarly to objects, we extract open-vocabulary features for rooms. In the feature extraction step (\cref{alg:open_vocab_low_level}), our system continuously computes CLIP embeddings ($\feature{\room}{\pose{}{}}$) of the full RGB frames ($\image{}{\text{RGB}}$) and associates each embedding with the agent’s corresponding pose ($\pose{}{}$) (\cref{low_level:full_clip}). In the room feature assignment module (\cref{alg:open_vocab_high_level}; see also Optimization \& Rooms in~\cref{fig:overview}a), we associate all the currently computed full RGB embeddings $\{\feature{\room}{\pose{0}{}}, \ldots, \feature{\room}{\pose{n}{}}\}$ at timestep $n$, with rooms based on spatial containment. For each detected room (\cref{high_level:detect_rooms}), we collect all CLIP embeddings linked to poses that lie within that room’s boundaries (\cref{high_level:find_room,high_level:append_feature}). Since CLIP embeddings of a room can vary significantly across viewpoints, we cluster the collected embeddings into $K$ groups using K-Means (\cref{high_level:cluster}). This produces a set of open-vocabulary feature clusters $\bF_{\room^i}$ associated with the $i$-th room $\node{\room^i}$. By clustering embeddings, we assume that images with a similar view frustum produce similar CLIP embeddings. 

\mypar{Object-level Relationships} We enhance our graph representation by explicitly modeling relationships between objects, leveraging the expressive power of a \vlm. After detecting objects of the input RGB-D frame ($\image{}{}$) in our feature extraction module (see~\cref{fig:overview}a and~\cref{alg:open_vocab_low_level}), we extract visual features for each pair of detected objects using a \vlm's visual encoder ($\feature{\text{\vlm}}{}$) and store them in a dictionary ($\relations$) whose keys are the detected objects' indices (\cref{low_level:ids_relations_map}). Our framework is designed to be compatible with any \vlm, as all architectures include a visual encoder, a visual-language adaptor and an \llm. In our graph $\graph$, we attach the output of the visual encoder, while the visual-language adaptor and \llm~are used in the reasoning module (\cref{subsec:reasoning_module}). 

The input to $\feature{\text{\vlm}}{}$ is the cropped region defined by the union of the two objects’ bounding boxes. Within this region, the boxes are inpainted into the image using a unique color for each label (see~\cref{fig:color_prompts}) to explicitly inform the \vlm~about the important objects to reason about. These features can later be combined with a text prompt to describe relationships between objects in the graph as natural language.  

When reconstructing the mesh (\cref{mid_level:meshing}, \cref{alg:open_vocab_mid_level}), we attach the detected object indices (called IDs) to the mesh, which are later used to assign an ID for each object node (\cref{mid_level:n_plus_1,mid_level:node_to_graph}). In~\cref{alg:open_vocab_rels_assignment}, we use these IDs to assign or update object-level relationships. If an edge between the objects already exists in the graph, we update its relationship feature using the following running average (\cref{rels:edge_in_graph,rels:runing_average_reasoning,rels:n_plus_1_relationship}):
\begin{align}
    \feature{\relationship}{i,j} = \dfrac{n_{\relationship}^{i,j}\feature{\relationship}{i,j} + \feature{\text{VLM}}{i,j}}{n_{\relationship}^{i,j} + 1}, \quad \feature{\relationship}{i,j}, n_{\relationship}^{i,j} \in \edge{\object^i}{\object^j}.
    \label{eq:average_relation}
\end{align}

\vspace{-0.5em}
Otherwise, we create a new edge containing the visual relationship feature (\cref{rels:edge_to_graph}). Finally, we remove the IDs from the mesh ($\nodes{\mesh}$), so that they do not interfere with the next iteration when assigning relationship features (\cref{rels_assignment:clean_mesh}).

\vspace{-0.5em}
\subsection{Task Reasoning}\label{subsec:reasoning_module}

In this section, we present a method that leverages the open-vocabulary semantics and object relationship information encoded in $\graph$ to reason about tasks. The goal is to assess task feasibility and generate a high-level plan that can guide downstream planning and scene interaction methods. Given a task in natural language, our approach consists of a three-stage reasoning pipeline as shown in~\cref{fig:overview}b.

\begin{figure}[t]
    \centering
    \includegraphics[width=\columnwidth]{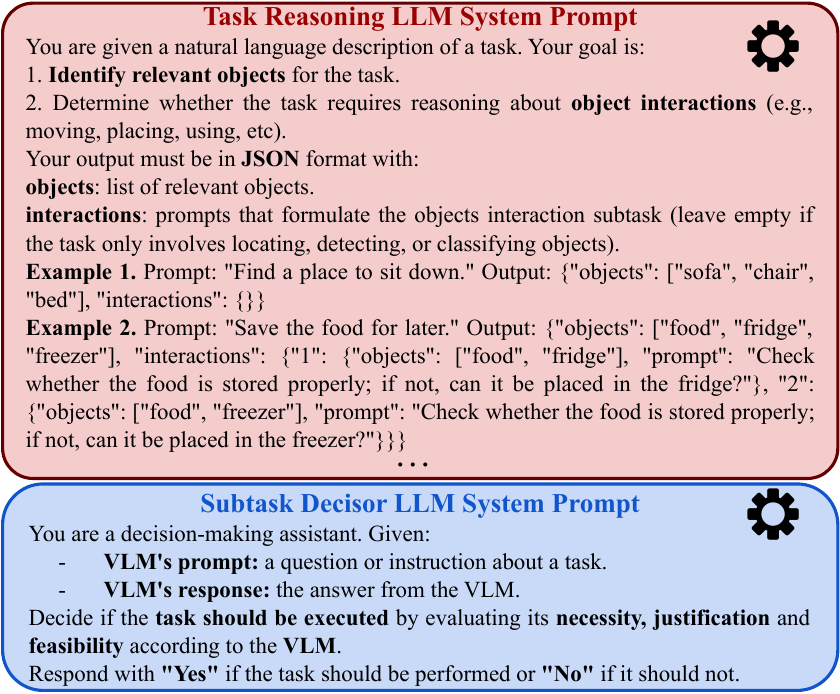}
    \caption{System prompts for task reasoning and subtask decisor \llm s.}
    \vspace{-0.5em}
\label{fig:prompts}
\end{figure}

\mypar{Task Reasoning} 
We use an \llm~to parse the input task. The \llm~identifies a list of $N_{\text{LLM}} \in \naturals$ task-relevant objects and, if the task requires interactions between these objects, it also enumerates those interactions framed as subtasks. For each subtask, the \llm~generates a natural language prompt to assess its feasibility. To ensure consistent and structured output, we design a system prompt, shown in~\cref{fig:prompts}, that guides the \llm~to follow this reasoning process and return a single JSON file containing the relevant objects, their subtasks (if required), and corresponding prompts. We further guide the \llm~by providing examples of tasks (\cref{fig:prompts} bottom part).

\mypar{Object and Room Search}
Given the list of $N_{\text{LLM}}$ task-relevant objects from the task reasoning \llm, the next step is to locate these objects within $\graph$. To achieve this, we first compute the $N_{\text{LLM}}$ CLIP~\cite{Radford2021CLIP} embeddings of the provided object names. We then compare these embeddings to those of the objects already in the scene graph by computing the cosine similarity against their open-vocabulary features $\feature{\object}{}$.  An object is considered found if its similarity exceeds a predefined threshold. If the task reasoning \llm~specifies subtasks between objects, we verify - after matching objects - that a relationship edge exists between the corresponding nodes in the graph. This relationship information is required for the subtask reasoning step performed by the \vlm. While objects in $\graph$ are associated with semantic labels, we rely on their CLIP embeddings for object search to mitigate label classification errors from the object detector.

Additionally, our method supports focusing on objects in specific rooms. To do this, we compute the cosine similarity between the CLIP embeddings of object names provided by the task reasoning \llm~and the embeddings of objects in each of the detected rooms. For each room, we average its $K  \times N_{\text{LLM}}$ cosine similarities and select rooms whose average exceeds a given threshold. This allows us to narrow down the object search and reasoning to relevant regions.

\setlength{\tabcolsep}{0.8pt}
\begin{table*}[t]
    \centering
    \begin{tabular}{l|cccccccc|cccccccc}
        \toprule
        & \multicolumn{8}{c|}{\textbf{HM3DSem}~\cite{Yadav2023HM3DSem}} & \multicolumn{8}{c}{\textbf{Replica}~\cite{replica19arxiv}}\\
        & $\text{Acc}_{5}$ & $\text{Acc}_{10}$ & $\text{Acc}_{25}$ & $\text{Acc}_{100}$ & $\text{Acc}_{250}$ & $\text{Acc}_{500}$ & \auc & \#objects & $\text{Acc}_{5}$ & $\text{Acc}_{10}$ & $\text{Acc}_{25}$ & $\text{Acc}_{100}$ & $\text{Acc}_{250}$ & $\text{Acc}_{500}$ & \auc & \#objects \\
        \midrule
        ConceptGraphs~\cite{Gu2024conceptgraphs} & \boldgreen{18.5} & \boldgreen{24.70} & \boldgreen{38.05} & \boldgreen{60.29} & \boldgreen{76.58} & \boldgreen{85.98} & \boldgreen{87.27} & \boldgreen{237.78} & \boldgreen{20.70} & \boldgreen{30.02} & \boldgreen{41.72} & \boldgreen{57.47} & \boldgreen{72.85} & \boldgreen{79.94} & \boldgreen{80.50} & \boldgreen{40.5} \\
        HOV-SG~\cite{werby2023hovsg} & 16.09 & 20.98 & 32.62 & 56.78 & 71.79 & 81.98 & 84.46 & \boldblue{405.33} & 3.66 & 6.56 & 11.26 & 24.00 & 43.65 & 63.27 & 66.82 & \boldblue{72.63} \\
        \midrule
        \method~(Ours) & \boldblue{37.72} & \boldblue{45.18} & \boldblue{56.52} & \boldblue{74.47} & \boldblue{85.51} & \boldblue{92.95} & \boldblue{92.49} & 88.33 & \boldblue{53.99} & \boldblue{58.23} & \boldblue{63.16} & \boldblue{73.74} & \boldblue{87.29} & \boldblue{88.80} & \boldblue{87.23} & 13.25 \\
        \bottomrule
    \end{tabular}
    \caption{Object retrieval quantitative results. \boldblue{Best result}, \boldgreen{Second-best result}. \method~outperforms baselines across metrics and datasets. Note that ConceptGraphs~\cite{Gu2024conceptgraphs} and HOV-SG~\cite{werby2023hovsg} operate on fully open-vocabulary segments, computed by clustering the embeddings of their point clouds.}
    \label{tab:object_retrieval_table}
\vspace{-2em}
\end{table*}

\begin{figure}[t]
    \centering
    \includegraphics[width=\columnwidth]{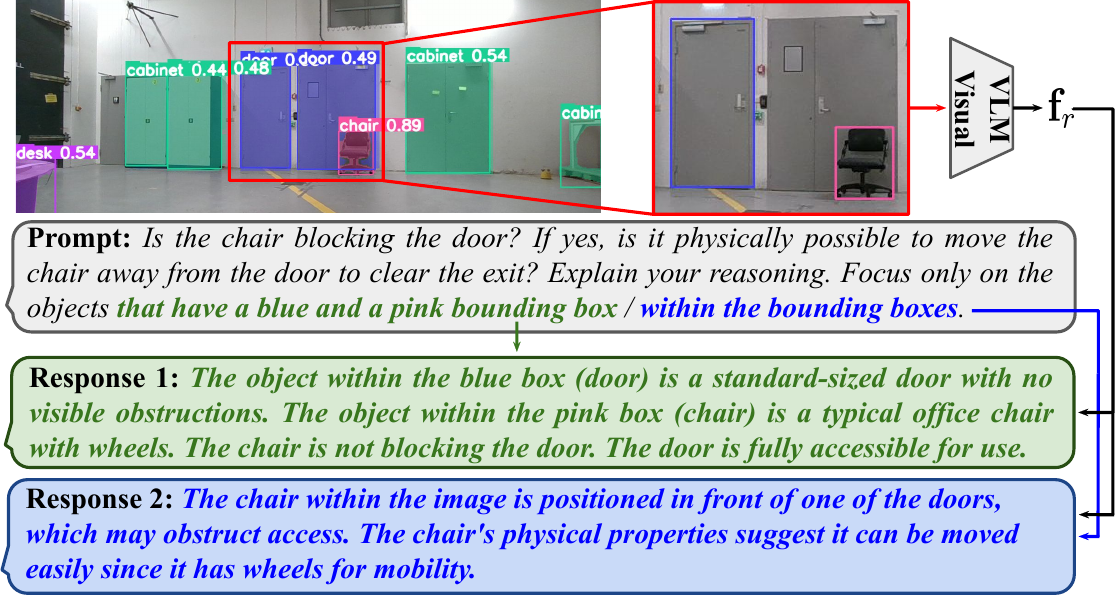}
    \caption{Guiding the \vlm~with bounding boxes colors. The \vlm~is provided with an image with a pair of objects and their inpainted bounding boxes, along with the task reasoning \llm~subtask prompt. In this example, the task is to reason about a door and a chair. Without including the bounding box colors in the prompt (\textcolor{blue}{blue}), the \vlm~focuses on the wrong object, reasoning about the door behind the chair. When the color information is included (\textcolor{Green}{green}), the \vlm~is guided to attend to the relevant objects.}
\label{fig:color_prompts}
\vspace{-0.5em}
\end{figure}

\mypar{VLM Reasoning}
If the task requires reasoning about interactions between objects, and the relevant objects have already been located in $\graph$, we use a \vlm~to evaluate whether each subtask is both necessary and feasible. We combine the subtasks generated by the task reasoning \llm~with the object-level relationship features ($\feature{\relationship}{}$) stored in $\graph$. In addition to the subtask generated by the task reasoning \llm, we append an additional prompt to guide the \vlm~focus on the correct objects in the image. Specifically, this hand-crafted prompt is: \textit{Focus only on the objects that have a X and Y bounding box}, where \textit{X} and \textit{Y} are the colors of the bounding boxes, inferred from the objects' labels $s_{\object}$. For this to work, each label in the set has a unique color associated with it. As shown in~\cref{fig:color_prompts}, providing this additional information helps the \vlm~attend to the objects of interest. In contrast, if the prompt only specifies generic bounding-boxed objects, the \vlm~may struggle to focus on the intended objects. 

Finally, a second \llm~interprets the \vlm's output to decide whether the subtasks should be executed. This \llm~is guided by a dedicated system prompt (see~\cref{fig:prompts}).

\section{EXPERIMENTS}\label{sec:results}

In this section, we first present the implementation details of our method (\cref{subsec:impl_details}). We then evaluate \method’s ability to construct scene graphs enriched with object-level open-vocabulary features (\cref{subsec:object_retrieval}). Next, we assess the task reasoning module in diverse real-world scenarios, using a quadruped robot and data collected by a human operator (\cref{subsec:task_reasoning}). Finally, we report the runtime performance of our method (\cref{subsec:runtime_eval}).

\vspace{-0.5em}
\subsection{Implementation Details}
\label{subsec:impl_details}
We employ YOLOe~\cite{wang2025yoloe} as our object detector, which outputs both bounding boxes and segmentation masks. For open-vocabulary feature extraction, we use CLIP~\cite{Radford2021CLIP}, specifically OpenAI’s ViT-L/14 model, chosen for its strong generalization across image classification tasks. We select Deepseek VL2~\cite{Lu2024DeepSeekVLTR} as our \vlm~due to its \sota performance. In addition, we use OpenAI’s o3 \llm~for task reasoning and parsing, and GPT-4o as subtask decisor \llm. 

Our sensor and compute setup consists of a Realsense D455 RGB-D camera, an Ouster OS0 LiDAR, and a VectorNav VN100 IMU for odometry, integrated with an NVIDIA Jetson Orin AGX for onboard processing (see~\cref{fig:example}). On the Orin AGX, we run the feature extraction modules (YOLOe, CLIP, and the VL2 visual encoder) as well as scene graph construction online at 1 Hz, while the room detection, mesh refinement and pose optimization are run at 0.5 Hz. The LLMs and the language components of VL2 are hosted in the cloud and queried only when the task is given. 

\begin{figure*}[t]
    \centering
    \includegraphics[width=1\textwidth]{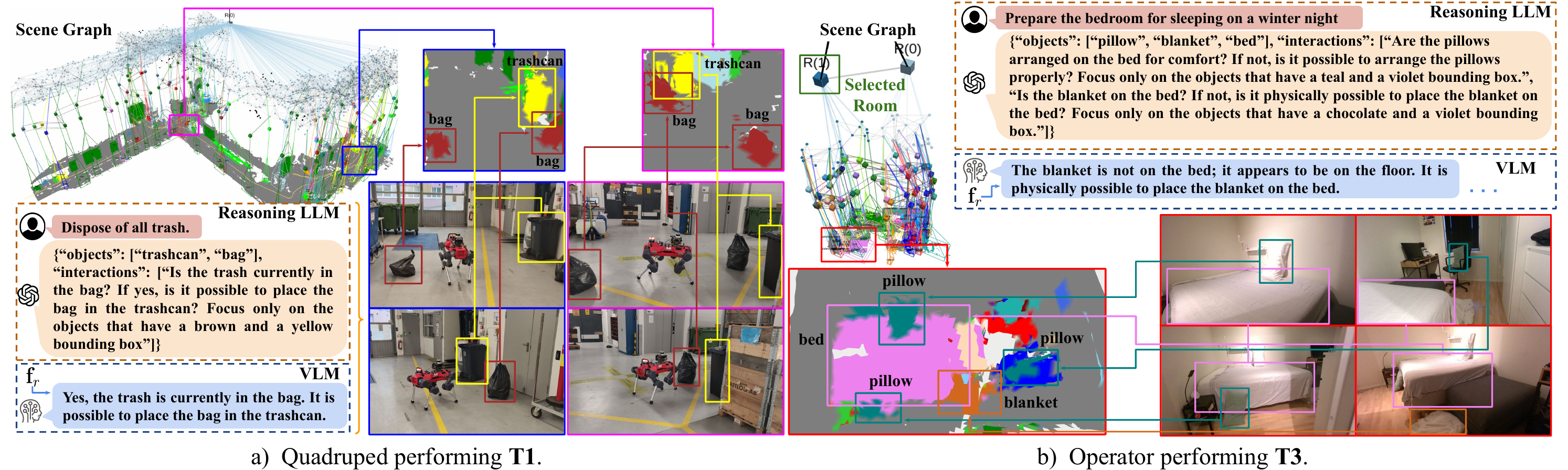}
    \caption{\method~performing \task{1} and \task{3}. The scene graph is built during exploration, after which the \llm~reasons about the task and the \vlm~evaluates subtasks using object relations ($\feature{\relationship}{}$). In both individual experiments, a 100 \sr~is achieved. In \task{3}, we apply our room search method, achieving 100\% accuracy in all 5 evaluations. We present one \vlm~reasoning example per task, although in practice the \vlm~reasons about each subtask.}
    \label{fig:examples}
\vspace{-1.5em}
\end{figure*}

\subsection{Open-Vocabulary Object Retrieval}
\label{subsec:object_retrieval}
To evaluate open-vocabulary object features, we compute the cosine similarity between each object feature and the dataset labels. We then rank the labels by similarity and measure the accuracy at the top-$k$ ranked predictions ($\text{Acc}_{k}$). Following HOV-SG~\cite{werby2023hovsg}, we also compute the area under the top-$k$ accuracy curve (\auc), which captures the alignment between predicted and ground-truth object categories across different values of $k$. We compare our approach against two strong open-vocabulary scene graph baselines, HOV-SG~\cite{werby2023hovsg} and ConceptGraphs~\cite{Gu2024conceptgraphs}, on the Replica~\cite{replica19arxiv} and Habitat Semantics (HM3DSem)~\cite{Yadav2023HM3DSem} datasets.

Both ConceptGraphs~\cite{Gu2024conceptgraphs} and HOV-SG~\cite{werby2023hovsg} construct scene graphs from fully open-vocabulary object segments, obtained by clustering CLIP embeddings on point clouds, which introduces background objects in their graphs. To ensure a fair comparison, we filter these out, as our method only includes foreground objects. Background objects generally degrade performance because their embeddings are more ambiguous due to a lack of texture and contamination from foreground objects. Specifically, we compute the cosine similarity between each detected object and a set of background categories (\eg~``wall'', ``floor'', ``ceiling'', ``stairs''), and discard objects with high similarity. Results in \cref{tab:object_retrieval_table} report top-$k$ accuracies ($\text{Acc}_{k}$) and \auc~scores for both our method and the baselines. \method~achieves substantially higher performance, showing its ability to assign correct open-vocabulary features to objects. Note that baselines detect a larger number of objects, since their fully open-vocabulary design, which detects objects by clustering on the embedding space, can fragment a single object into multiple segments when CLIP embeddings vary across viewpoints of the same object. We nevertheless include these methods as baselines because they represent the current \sota in open-vocabulary scene graph construction.

\subsection{Real-World Task Reasoning Evaluation}
\label{subsec:task_reasoning}

To assess the reasoning capabilities of \method~in handling complex tasks, we design a series of evaluation tasks that require identifying objects within the graph and, in some cases, reasoning about their relations. Since no dedicated evaluation dataset exists, we choose to conduct real experiments, which not only provide the necessary evaluation data but also allow us to verify the robustness of our method in practice. Performance is measured using two metrics: the success ratio (\sr), defined as the percentage of correctly evaluated subtasks or successfully identified objects, and the number of false positives (\fp), defined as the number of incorrectly evaluated subtasks or misidentified objects. For each task, we manually identify the ground truth objects or object pairs whose subtask should be positively evaluated. Specifically, we consider the following tasks:

\begin{itemize}
\item[\task{1:}] \textbf{Dispose of all trash}: Identify filled trash bags near trash cans and determine if they can be thrown away.
\item[\task{2:}] \textbf{Ensure exits and entrances are not blocked}: Detect objects that may be blocking doorways.
\item[\task{3:}] \textbf{Prepare the bedroom for sleeping on a winter night}: Verify whether the pillows and blankets are placed appropriately on the bed.
\item[\task{4:}] \textbf{Prepare the meeting room for floor cleaning}: Determine whether chairs and other objects can be placed on desks to facilitate cleaning.
\item[\task{5:}] \textbf{Find a backpack, a fan, plants and trash cans.} 
\end{itemize}

For \task{1}, \task{2}, and \task{5}, \method~is deployed on a quadruped robot equipped with the sensing and computing module introduced in~\cref{subsec:impl_details}. The robot autonomously explores the environment using a graph-based path planning approach~\cite{dang2020gbplanner}, while incrementally constructing the scene graph. The same planner~\cite{dang2020gbplanner} is used to navigate to detected objects or object pairs of positively evaluated subtasks. In contrast, for \task{3} and \task{4}, data is collected with the same sensing module, but carried by a human operator.

\begin{figure}[t]
    \centering
    \includegraphics[width=\columnwidth]{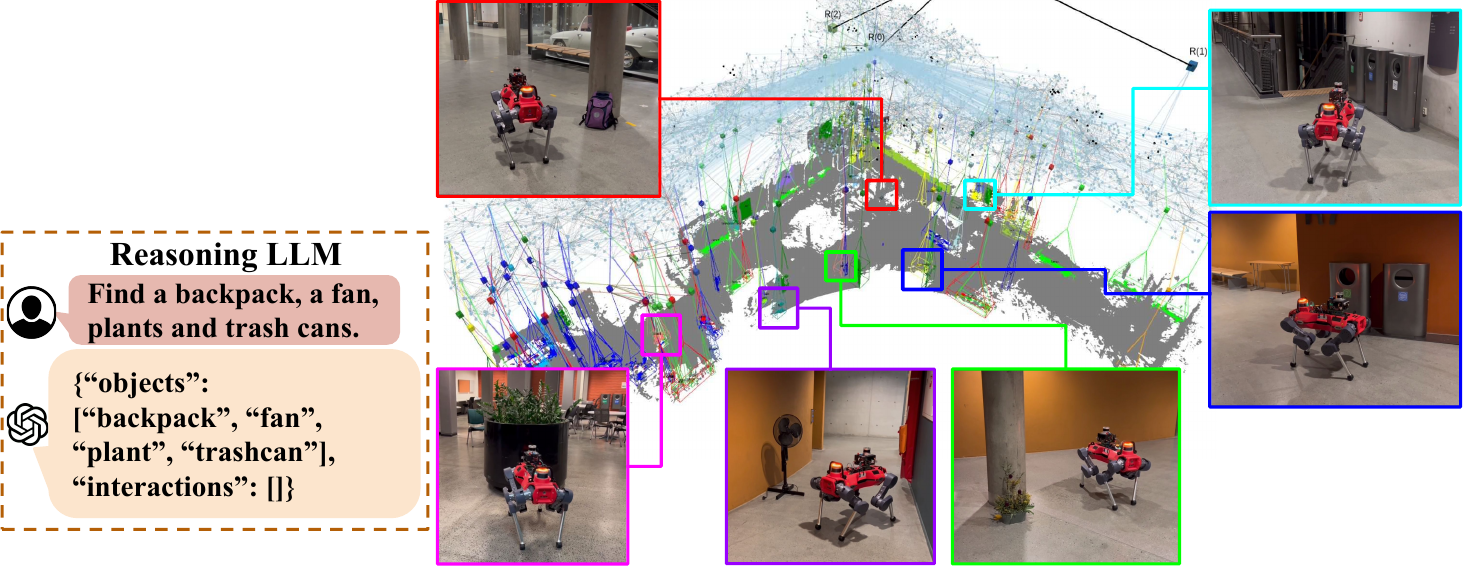}
    \caption{Object search task (\task{5}). After the scene graph is constructed during autonomous exploration, we ask \method~to find a backpack, a fan, plants and trash cans. Our method is capable of finding all the objects.}
\label{fig:object_search_example}
\vspace{-.5em}
\end{figure}

Once the scene graph is constructed, task reasoning begins. As detailed in~\cref{subsec:reasoning_module}, the task reasoning \llm~identifies task-relevant objects, which are then located in the scene graph using CLIP-based cosine similarity. In \task{5}, illustrated in~\cref{fig:object_search_example}, this retrieval step is sufficient, as no reasoning about object interactions is required. For the remaining tasks, the \vlm~is prompted with the object relationships ($\feature{\relationship}{}$) and the subtask prompts generated by the task reasoning~\llm.

\cref{tab:tasks} presents the success ratio and number of false positives with standard deviation across five evaluations for each task, comparing two VLMs: DeepSeek VL2~\cite{Lu2024DeepSeekVLTR} and InstructBLIP~\cite{dai2023instructblip}. \method~(with VL2) achieves consistently high \sr~(84-100) with low FP across all tasks. DeepSeek VL2 consistently outperforms InstructBLIP, achieving higher success ratios and fewer false positives, supporting its selection as the primary VLM in \method. \#GT+ denotes the number of positive subtasks in each task. \cref{fig:examples}a illustrates \task{1} on the quadruped robot, where the system reasons about placing four trash bags inside nearby trashcans, and \cref{fig:examples}b illustrates \task{3}, involving arranging blankets and pillows on a bed. \method~achieves high \sr~in both examples.

To further assess the performance of the object-level relationships ($\feature{\relationship}{}$), the \vlm~(VL2), and the subtask decisor \llm, we use the collected data from tasks \task{1} to \task{4} to re-evaluate all subtasks 100 times, accounting for the stochasticity of both models. As shown in~\cref{tab:vlm_metrics}, the combined system achieves consistently high accuracy, defined as the percentage of correctly evaluated subtasks, across all tasks. The macro-averaged $F_1$ scores, which balance precision and recall, indicate strong performance on both positive and negative subtasks, \ie~those that should and should not be executed. The pooled results highlight the robustness of our method, confirming its ability to reason about object interactions despite the stochasticity of the \vlm~and \llm.

\setlength{\tabcolsep}{1.5pt}
\begin{table}[t]
    \centering
    \begin{tabular}{l|c|ccccc}
        \toprule
         & VLM & \task{1} & \task{2} & \task{3} & \task{4} & \task{5} \\
        \midrule
        \multirow{2}{*}{\sr} & VL2~\cite{Lu2024DeepSeekVLTR} & 90 $\pm$ 12.3 & 84 $\pm$ 15.0 & 90 $\pm$ 12.3 & 86 $\pm$ 7.1 & \multirow{2}{*}{100} \\
                & BLIP~\cite{dai2023instructblip} & 55 $\pm$ 24.5 & 48 $\pm$ 9.8 & 40 $\pm$ 12.2 & 53.3 $\pm$ 12.5 & \\
        \midrule
        \multirow{2}{*}{FP}      & VL2~\cite{Lu2024DeepSeekVLTR} & 0.2 $\pm$ 0.4 & 0.8 $\pm$ 0.4 & 0.6 $\pm$ 0.5 & 0.8 $\pm$ 0.4 & \multirow{2}{*}{1.8 $\pm$ 0.4} \\
                & BLIP~\cite{dai2023instructblip} & 0.2 $\pm$ 0.4 & 1.8 $\pm$ 0.4 & 0.6 $\pm$ 0.49 & 1.0 $\pm$ 0.0 &  \\
        \midrule
        \multicolumn{2}{c|}{\#GT+}  & 4 & 5 & 4 & 6 & 7 \\
        \bottomrule
    \end{tabular}
    \caption{Comparison of task reasoning performance using two VLMs (DeepSeek VL2~\cite{Lu2024DeepSeekVLTR} and InstructBLIP~\cite{dai2023instructblip}). \task{1} - \task{4} require relational reasoning, while \task{5} does not. Each task is executed 5 times, and the results are averaged. Our method with VL2 achieves higher success ratios with fewer false positives, motivating its choice as the primary VLM.}
    \label{tab:tasks}
    \vspace{-0.6em}
\end{table}

\setlength{\tabcolsep}{4pt}
\begin{table}[t]
    \centering
    \begin{tabular}{l|ccccc}
        \toprule
        Tasks & \task{1} & \task{2} & \task{3} & \task{4} & Pooled \\
        \midrule
        Accuracy (\%)              & 88.81 & 84.43 & 83.11 & 74.91 & 79.30 \\
        $F_1$ (\%)                 & 88.55 & 85.43 & 81.86 & 75.84 & 79.11 \\
        Average FP                     & 0.27 & 1.04 & 0.69 & 1.12 & 0.78 \\
        \midrule
        \# Positive subtasks       & 18 & 21 & 20 & 44 & 103 \\
        \# Negative subtasks       & 3 & 75 & 7 & 11 & 96 \\
        \bottomrule
    \end{tabular}
    \caption{\vlm~and subtask decisor \llm~performance. We evaluate each subtask from \task{1} - \task{4} 100 times. This accounts for the stochasticity introduced by both the \vlm~and the decisor \llm.}
    \label{tab:vlm_metrics}
    \vspace{-.6em}
    
\end{table}

\setlength{\tabcolsep}{3.5pt}
\begin{table}[t]
    \centering
    \begin{tabular}{l|cccc}
        \toprule
        Tasks & YOLOe \cite{wang2025yoloe} & CLIP \cite{Radford2021CLIP} & VL2 Visual \cite{Lu2024DeepSeekVLTR} & Graph \\
        \midrule
        \task{1} & 44.9 $\pm$ 29.5 & 76.1 $\pm$ 56.6 & 214.8 $\pm$ 147 & 262.2 $\pm$ 105.5 \\
        \task{2} & 54.9 $\pm$ 43 & 97.4 $\pm$ 59 & 257.8 $\pm$ 132.4 & 231.1 $\pm$ 86.3 \\
        \task{3} & 70.7 $\pm$ 63.6 & 133.1 $\pm$ 95.5 & 277.8 $\pm$ 141.2 & 144.9 $\pm$ 91.8 \\
        \task{4} & 87.9 $\pm$ 33.8 & 157.9 $\pm$ 69.1 & 223.8 $\pm$ 145.3 & 213.5 $\pm$ 84.2 \\
        \task{5} & 54.8 $\pm$ 57.2 & 82.9 $\pm$ 66.9 & 247 $\pm$ 140.8 & 276 $\pm$ 92.2 \\  
        \bottomrule
    \end{tabular}
    \caption{Timing statistics (mean $\pm$ standard deviation) in milliseconds.}
    \label{tab:timings}
    \vspace{-.7em}
\end{table}

\vspace{-0.5em}
\subsection{Runtime Evaluation}
\label{subsec:runtime_eval}
To verify the feasibility of running our graph construction online, we measure the runtime of its main components, as reported in Table~\ref{tab:timings}. On average, YOLOe and CLIP inference require less than 160 ms per frame, while the VL2 visual encoder is the most expensive component, ranging from 214 - 278 ms. Scene graph construction remains lightweight, with runtimes between 145 - 276 ms. Overall, the method operates below one second per frame. This runtime supports online operation on the Orin AGX mounted on the quadruped robot, since its speeds allow running the method at 1 - 2 Hz without skipping important information.

\vspace{-1.5em}
\section{CONCLUSIONS}\label{sec:conclusions}

We propose \method, a framework for constructing hierarchical 3D scene graphs that incorporate open-vocabulary features and encode object-level relationships using a \vlm. Our reasoning module further leverages both \llm s and a \vlm~to interpret complex tasks, identify task-relevant objects, and reason about their potential interactions. Extensive experiments show that \method~outperforms existing open-vocabulary scene graph baselines in object retrieval and consistently solves diverse reasoning tasks with high success rates and low false positives. Deployment on a quadruped robot demonstrates the framework’s capability for online, incremental scene understanding and task reasoning. These results demonstrate the potential of combining hierarchical representations with vision-language reasoning to achieve richer context-aware understanding.

\bibliographystyle{IEEEtran}
\bibliography{IEEEabrv, bib_short2}

\end{document}